\documentclass{article}

\usepackage[final]{neurips_2019}

\usepackage[utf8]{inputenc} %
\usepackage[T1]{fontenc}    %
\usepackage{hyperref}       %
\usepackage{url}            %
\usepackage{booktabs}       %
\usepackage{amsfonts}       %
\usepackage{nicefrac}       %
\usepackage{microtype}      %
\usepackage{csquotes}       %
\usepackage{makecell}       %
\usepackage{lipsum}
\usepackage{graphicx}
\usepackage[toc,page]{appendix}
\usepackage{amsmath}
\usepackage{float}
\usepackage{caption} 
\usepackage[english]{babel}%
\usepackage[style=numeric, doi=false, isbn=false, url=false, eprint=false, backend=bibtex]{biblatex}
\title{Recurrent Neural Networks (RNNs):\\ A gentle Introduction and Overview}

\author{%
  Robin M. Schmidt \\
  Department of Computer Science\\
  Eberhard-Karls-University T\"ubingen\\
  T\"ubingen, Germany \\
  \texttt{rob.schmidt@student.uni-tuebingen.de} \\
}

\begin{document}

\maketitle

\begin{abstract}
  State-of-the-art solutions in the areas of ``Language Modelling \& Generating Text'', ``Speech Recognition'', ``Generating Image Descriptions'' or ``Video Tagging'' have been using Recurrent Neural Networks as the foundation for their approaches. Understanding the underlying concepts is therefore of tremendous importance if we want to keep up with recent or upcoming publications in those areas. In this work we give a short overview over some of the most important concepts in the realm of Recurrent Neural Networks which enables readers to easily understand the fundamentals such as but not limited to ``Backpropagation through Time'' or ``Long Short-Term Memory Units'' as well as some of the more recent advances like the ``Attention Mechanism'' or ``Pointer Networks''. We also give recommendations for further reading regarding more complex topics where it is necessary.
\end{abstract}

\section{Introduction \& Notation}
\label{sec:introduction}
\begin{figure}
    \centering
    \includegraphics[width=\textwidth]{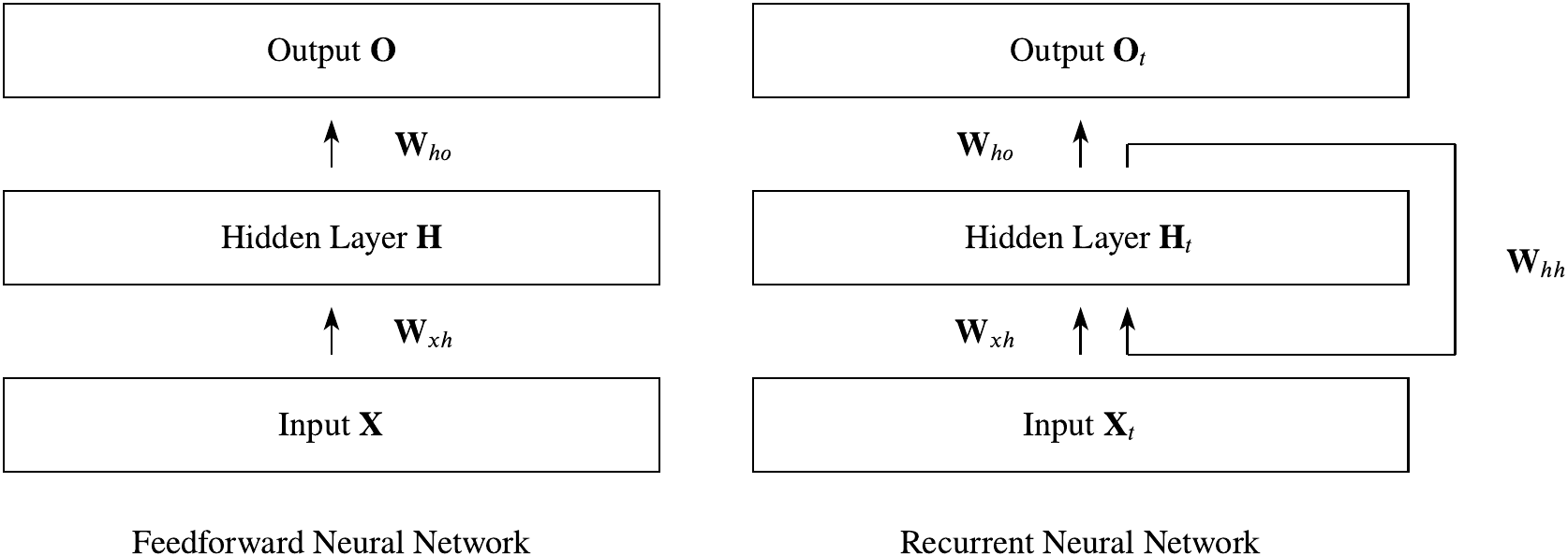}
    \caption{Visualisation of differences between Feedfoward NNs und Recurrent NNs}
    \label{fig:rnn}
\end{figure}

Recurrent Neural Networks (RNNs) are a type of neural network architecture which is mainly used to detect patterns in a sequence of data. Such data can be handwriting, genomes, text or numerical time series which are often produced in industry settings (e.g. stock markets or sensors) \cite{chung2015gated, bworld}. However, they are also applicable to images if these get respectively decomposed into a series of patches and treated as a sequence \cite{bworld}. On a higher level, RNNs find applications in \textit{Language Modelling \& Generating Text}, \textit{Speech Recognition}, \textit{Generating Image Descriptions} or \textit{Video Tagging}. What differentiates Recurrent Neural Networks from Feedforward Neural Networks also known as Multi-Layer Perceptrons (MLPs) is how information gets passed through the network. While Feedforward Networks pass information through the network without cycles, the RNN has cycles and transmits information back into itself. This enables them to extend the functionality of Feedforward Networks to also take into account previous inputs $\mathbf{X}_{0:t-1}$ and not only the current input $\mathbf{X}_t$. This difference is visualised on a high level in Figure \ref{fig:rnn}. Note, that here the option of having multiple hidden layers is aggregated to one Hidden Layer block $\mathbf{H}$. This block can obviously be extended to multiple hidden layers.

We can describe this process of passing information from the previous iteration to the hidden layer with the mathematical notation proposed in \cite{zhang2019dive}. For that, we denote the hidden state and the input at time step $t$ respecively as $\mathbf{H}_{t} \in \mathbb{R}^{n \times h}$ and $\mathbf{X}_t \in \mathbb{R}^{n \times d}$ where $n$ is number of samples, $d$ is the number of inputs of each sample and $h$ is the number of hidden units. Further, we use a weight matrix $\mathbf{W}_{x h} \in \mathbb{R}^{d \times h}$, hidden-state-to-hidden-state matrix $\mathbf{W}_{h h} \in \mathbb{R}^{h \times h}$ and a bias parameter $\mathbf{b}_{h} \in \mathbb{R}^{1 \times h}$. Lastly, all these informations get passed to a activation function $\phi$ which is usually a logistic sigmoid or tanh function to prepair the gradients for usage in backpropagation. Putting all these notations together yields Equation \ref{eq:not} as the hidden variable and Equation \ref{eq:outrnn} as the output variable.
\begin{equation}
\label{eq:not}
    \mathbf{H}_{t}=\phi_h\left(\mathbf{X}_{t} \mathbf{W}_{x h}+\mathbf{H}_{t-1} \mathbf{W}_{h h}+\mathbf{b}_{h}\right)
\end{equation}
\begin{equation}
\label{eq:outrnn}
    \mathbf{O}_{t}=\phi_o\left(\mathbf{H}_{t} \mathbf{W}_{h o}+\mathbf{b}_{o}\right)
\end{equation}

Since $\mathbf{H}_{t}$ recursively includes $\mathbf{H}_{t-1}$ and this process occurs for every time step the RNN includes traces of all hidden states that preceded $\mathbf{H}_{t-1}$ as well as $\mathbf{H}_{t-1}$ itself.

If we compare that notation for RNNs with similar notation for Feedforward Neural Networks we can clearly see the difference we described earlier. In Equation \ref{eq:Hidden} we can see the computation for the hidden variable while Equation \ref{eq:output} shows the output variable.
\begin{equation}
\label{eq:Hidden}
    \mathbf{H} =\phi_h\left(\mathbf{X} \mathbf{W}_{x h}+\mathbf{b}_{h}\right)
\end{equation}
\begin{equation}
\label{eq:output}
    \mathbf{O}=\phi_o \left(\mathbf{H W}_{h o}+\mathbf{b}_{o}\right)
\end{equation}

If you are familiar with training techniques for Feedforward Neural Networks such as backpropagation one question which might arise is how to properly backpropagate the error through a RNN. Here, a technique called Backpropagation Through Time (BPTT) is used which gets described in detail in the next section.

\section{Backpropagation Through Time (BPTT) \& Truncated BPTT}
Backpropagation Through Time (BPTT) is the adaption of the backpropagation algorithm for RNNs \cite{zhang2019dive}. In theory, this unfolds the RNN to construct a traditional Feedfoward Neural Network where we can apply backpropagation. For that, we use the same notations for the RNN as proposed before.

When we forward pass our input $\mathbf{X}_t$ through the network we compute the hidden state $\mathbf{H}_t$ and the output state $\mathbf{O}_t$ one step at a time. We can then define a loss function $\mathcal{L}\left(\mathbf{O}, \mathbf{Y}\right)$ to describe the difference between all outputs $\mathbf{O}_t$ and target values $\mathbf{Y}_t$ as shown in Equation \ref{eq:loss}. This basically sums up every loss term $\ell_t$ of each update step so far. This loss term $\ell_t$ can have different definitions based on the specific problem (e.g. Mean Squared Error, Hinge Loss, Cross Entropy Loss, etc.).
\begin{equation}
\label{eq:loss}
    \mathcal{L}\left(\mathbf{O}, \mathbf{Y}\right)=\sum_{t=1}^{T} \ell_t\left(\mathbf{O}_{t}, \mathbf{Y}_{t}\right)
\end{equation}
Since we have three weight matrices $\mathbf{W}_{x h}$, $\mathbf{W}_{h h}$ and $\mathbf{W}_{h o}$ we need to compute the partial derivative w.r.t. to each of these weight matrices. With the chain rule which is also used in normal backpropagation we get to the result for $\mathbf{W}_{h o}$ shown in Equation \ref{eq:derive_ho}.
\begin{equation}
\label{eq:derive_ho}
    \frac{\partial \mathcal{L}}{\partial \mathbf{W}_{ho}} = \sum_{t=1}^{T} \frac{\partial \ell_t}{\partial \mathbf{O}_t} \cdot \frac{\partial \mathbf{O}_t}{\partial \phi_o} \cdot \frac{\partial \phi_o}{\mathbf{W}_{ho}} = \sum_{t=1}^{T} \frac{\partial \ell_t}{\partial \mathbf{O}_t} \cdot \frac{\partial \mathbf{O}_t}{\partial \phi_o} \cdot \mathbf{H_t}
\end{equation}

For the partial derivative with respect to $\mathbf{W}_{hh}$ we get the result shown in Equation \ref{eq:derive_hh}.
\begin{equation}
\label{eq:derive_hh}
    \frac{\partial \mathcal{L}}{\partial \mathbf{W}_{hh}} = \sum_{t=1}^{T} \frac{\partial \ell_t}{\partial \mathbf{O}_t} \cdot \frac{\partial \mathbf{O}_t}{\partial \phi_o} \cdot \frac{\partial \phi_o}{\partial \mathbf{H}_t} \cdot \frac{\partial \mathbf{H}_t}{\partial \phi_h} \cdot \frac{\partial \phi_h}{\partial \mathbf{W}_{hh}} = \sum_{t=1}^{T} \frac{\partial \ell_t}{\partial \mathbf{O}_t} \cdot \frac{\partial \mathbf{O}_t}{\partial \phi_o} \cdot \mathbf{W}_{ho} \cdot \frac{\partial \mathbf{H}_t}{\partial \phi_h} \cdot \frac{\partial \phi_h}{\partial \mathbf{W}_{hh}}
\end{equation}

For the partial derivative with respect to $\mathbf{W}_{xh}$ we get the result shown in Equation \ref{eq:derive_xh}.
\begin{equation}
\label{eq:derive_xh}
    \frac{\partial \mathcal{L}}{\partial \mathbf{W}_{xh}} = \sum_{t=1}^{T} \frac{\partial \ell_t}{\partial \mathbf{O}_t} \cdot \frac{\partial \mathbf{O}_t}{\partial \phi_o} \cdot \frac{\partial \phi_o}{\partial \mathbf{H}_t} \cdot \frac{\partial \mathbf{H}_t}{\partial \phi_h} \cdot \frac{\partial \phi_h}{\partial \mathbf{W}_{xh}} = \sum_{t=1}^{T} \frac{\partial \ell_t}{\partial \mathbf{O}_t} \cdot \frac{\partial \mathbf{O}_t}{\partial \phi_o} \cdot \mathbf{W}_{ho} \cdot \frac{\partial \mathbf{H}_t}{\partial \phi_h} \cdot \frac{\partial \phi_h}{\partial \mathbf{W}_{xh}}
\end{equation}

Since each $\mathbf{H}_t$ depends on the previous time step we can substitute the last part from above equations to get Equation \ref{trick_hh} and Equation \ref{trick_xh}.
\begin{equation}
\label{trick_hh}
     \frac{\partial \mathcal{L}}{\partial \mathbf{W}_{hh}} = \sum_{t=1}^{T} \frac{\partial \ell_t}{\partial \mathbf{O}_t} \cdot \frac{\partial \mathbf{O}_t}{\partial \phi_o} \cdot \mathbf{W}_{ho} \sum_{k=1}^t \frac{\partial \mathbf{H}_t}{\partial \mathbf{H}_k} \cdot \frac{\partial \mathbf{H}_k}{\partial \mathbf{W}_{hh}}
\end{equation}
\begin{equation}
\label{trick_xh}
     \frac{\partial \mathcal{L}}{\partial \mathbf{W}_{xh}} = \sum_{t=1}^{T} \frac{\partial \ell_t}{\partial \mathbf{O}_t} \cdot \frac{\partial \mathbf{O}_t}{\partial \phi_o} \cdot \mathbf{W}_{ho} \sum_{k=1}^t \frac{\partial \mathbf{H}_t}{\partial \mathbf{H}_k} \cdot \frac{\partial \mathbf{H}_k}{\partial \mathbf{W}_{xh}}
\end{equation}

The adapted part can then further be written as shown in Equation \ref{finish_hh} and Equation \ref{finish_xh}.
\begin{equation}
\label{finish_hh}
     \frac{\partial \mathcal{L}}{\partial \mathbf{W}_{hh}} = \sum_{t=1}^{T} \frac{\partial \ell_t}{\partial \mathbf{O}_t} \cdot \frac{\partial \mathbf{O}_t}{\partial \phi_o} \cdot \mathbf{W}_{ho} \sum_{k=1}^t \left( \mathbf{W}_{hh}^\top \right)^{t-k} \cdot \mathbf{H}_k
\end{equation}
\begin{equation}
\label{finish_xh}
     \frac{\partial \mathcal{L}}{\partial \mathbf{W}_{xh}} = \sum_{t=1}^{T} \frac{\partial \ell_t}{\partial \mathbf{O}_t} \cdot \frac{\partial \mathbf{O}_t}{\partial \phi_o} \cdot \mathbf{W}_{ho} \sum_{k=1}^t \left( \mathbf{W}_{hh}^\top \right)^{t-k} \cdot \mathbf{X}_k
\end{equation}

From here, we can see that we need to store powers of $\mathbf{W}_{hh}^k$ as we proceed through each loss term $\ell_t$ of the overall loss function $\mathcal{L}$ which can become very large. For these large values this method becomes numerically unstable since eigenvalues smaller than $1$ vanish and eigenvalues larger than $1$ diverge \cite{longterm}. One method of solving this problem is truncate the sum at
a computationally convenient size \cite{zhang2019dive}. When you do this, you're using Truncated BPTT \cite{trunc}. This basically establishes an upper bound for the number of time steps the gradient can flow back to \cite{Sutskever:2013:TRN:2604780}. One can think of this upper bound as a moving window of past time steps which the RNN considers. Anything before the cut-off time step doesn't get taken into account. Since BPTT basically unfolds the RNN to create a new layer for each time step we can also think of this procedure as limiting the number of hidden layers. 

\section{Problems of RNNs: Vanishing \& Exploding Gradients}
As in most neural networks, vanishing or exploding gradients is a key problem of RNNs \cite{bworld}. In Equation \ref{trick_hh} and Equation \ref{trick_xh} we can see $\frac{\partial \mathbf{H}_t}{\partial \mathbf{H}_k}$ which basically introduces matrix multiplication over the (potentially very long) sequence, if there are small values ($< 1$) in the matrix multiplication this causes the gradient to decrease with each layer (or time step) and finally vanish  \cite{chen2016gentle}. This basically stops the contribution of states that happened far earlier than the current time step towards the current time step  \cite{chen2016gentle}. Similarly, this can happen in the opposite direction if we have large values ($>1$) during matrix multiplication causing an exploding gradient which in result values each weight too much and changes it heavily \cite{chen2016gentle}.

This problem motivated the introduction of the long short term memory units (LSTMs) to particularly handle the vanishing gradient problem. This approach was able to outperform traditional RNNs on a variety of tasks \cite{chen2016gentle}. In the next section we want to go deeper on the proposed structure of LSTMs.

\section{Long Short-Term Memory Units (LSTMs)}
\label{sec:lstms}
Long Short-Term Memory Units (LSTMs) \cite{lstms} were designed to properly handle the vanishing gradient problem. Since they use a more constant error, they allow RNNs to learn over a lot more time steps (way over $1000$) \cite{bworld}. To achieve that, LSTMs store more information outside of the traditional neural network flow in structures called gated cells \cite{chen2016gentle, bworld}. To make things work in an LSTM we use an output gate $\mathbf{O}_{t}$ to read entries of the cell, an input gate $\mathbf{I}_{t}$ to read data into the cell and a forget gate $\mathbf{F}_{t}$ to reset the content of the cell. The computations for these gates are shown in Equation \ref{eq:output_gate}, Equation \ref{eq:input_gate} and Equation \ref{eq:forget_gate}. For a more visual approach please see Figure \ref{fig:gates} in Appendix \ref{app:vis}. 
\begin{equation}
\label{eq:output_gate}
\mathbf{O}_{t}=\sigma\left(\mathbf{X}_{t} \mathbf{W}_{x o}+\mathbf{H}_{t-1} \mathbf{W}_{h o}+\mathbf{b}_{o}\right)
\end{equation}
\begin{equation}
\label{eq:input_gate}
    \mathbf{I}_{t}=\sigma\left(\mathbf{X}_{t} \mathbf{W}_{x i}+\mathbf{H}_{t-1} \mathbf{W}_{h i}+\mathbf{b}_{i}\right)
\end{equation}
\begin{equation}
\label{eq:forget_gate}
\mathbf{F}_{t}=\sigma\left(\mathbf{X}_{t} \mathbf{W}_{x f}+\mathbf{H}_{t-1} \mathbf{W}_{h f}+\mathbf{b}_{f}\right)
\end{equation}

The shown equations use $\mathbf{W}_{x i}, \mathbf{W}_{x f}, \mathbf{W}_{x o} \in \mathbb{R}^{d \times h}$ and $\mathbf{W}_{h i}, \mathbf{W}_{h f}, \mathbf{W}_{h o} \in \mathbb{R}^{h \times h}$ as weight matrices while $\mathbf{b}_{i}, \mathbf{b}_{f}, \mathbf{b}_{o} \in \mathbb{R}^{1 \times h}$ are their respective biases. Further, they use the sigmoid activation function $\sigma$ to transform the output $\in (0,1)$ which each results in a vector with entries $\in (0,1)$. 

Next, we need a candidate memory cell $\tilde{\mathbf{C}}_{t} \in \mathbb{R}^{n \times h}$ which has a similar computation as the previously mentioned gates but instead uses a tanh activation function to have an output $\in (-1,1)$. Further, it again has its own weights $\mathbf{W}_{x c} \in \mathbb{R}^{d \times h}$, $\mathbf{W}_{h c} \in \mathbb{R}^{h \times h}$ and biases $\mathbf{b}_{c} \in \mathbb{R}^{1 \times h}$. The respective computation is shown in Equation \ref{eq:candidate}. See Figure \ref{fig:candidate} in Appendix \ref{app:vis} for a visualisation of this enhancement.
\begin{equation}
\label{eq:candidate}
\tilde{\mathbf{C}}_{t}=\tanh \left(\mathbf{X}_{t} \mathbf{W}_{x c}+\mathbf{H}_{t-1} \mathbf{W}_{h c}+\mathbf{b}_{c}\right)
\end{equation}

To plug some things together we introduce old memory content $\mathbf{C}_{t-1} \in \mathbb{R}^{n \times h}$ which together with the introduced gates controls how much of the old memory content we want to preserve to get to the new memory content $\mathbf{C}_t$. This is shown in Equation \ref{eq:memory} where $\odot$ denotes element-wise multiplication. The structure so far can be seen in Figure \ref{fig:memory} in Appendix \ref{app:vis}.
\begin{equation}
\label{eq:memory}
\mathbf{C}_{t}=\mathbf{F}_{t} \odot \mathbf{C}_{t-1}+\mathbf{I}_{t} \odot \tilde{\mathbf{C}}_{t}
\end{equation}

The last step is to introduce the computation for the hidden states $\mathbf{H}_{t} \in \mathbb{R}^{n \times h}$ into the framework. This can be seen in Equation \ref{eq:lstm_hidden}.
\begin{equation}
\label{eq:lstm_hidden}
\mathbf{H}_{t}=\mathbf{O}_{t} \odot \tanh \left(\mathbf{C}_{t}\right)
\end{equation}
With the tanh function we ensure  that each element of $\mathbf{H}_{t}$ is $\in (-1,1)$. The full LSTM framework can be seen in Figure \ref{fig:hidden} in Appendix \ref{app:vis}.

\section{Deep Recurrent Neural Networks (DRNNs)}
Deep Recurrent Neural Networks (DRNNs) are in theory a really easy concept. To construct a deep RNN with $L$ hidden layers we simply stack ordinary RNNs of any type on top of each other. Each hidden state $\mathbf{H}_t^{(\ell)} \in \mathbb{R}^{n \times h}$ is passed to the next time step of the current layer $\mathbf{H}_{t+1}^{(\ell)}$ as well as the current time step of the next layer $\mathbf{H}_{t}^{(\ell+1)}$. For the first layer we compute the hidden state as proposed in the previous models shown shown in Equation \ref{eq:layer1} while for the subsequent layer we use Equation \ref{eq:conslayer} where the hidden state from the previous layer is treated as input. 
\begin{equation}
\label{eq:layer1}
\mathbf{H}_{t}^{(1)}=\phi_{1}\left(\mathbf{X}_{t}, \mathbf{H}_{t-1}^{(1)}\right)
\end{equation}
\begin{equation}
\label{eq:conslayer}
\mathbf{H}_{t}^{(\ell)}=\phi_{\ell}\left(\mathbf{H}_{t}^{(\ell-1)}, \mathbf{H}_{t-1}^{(\ell)}\right)
\end{equation}

The output $\mathbf{O}_t \in \mathbb{R}^{n \times o}$ where $o$ is the number of outputs is then computed as shown in Equation \ref{outputdnn} where we only use the hidden state of layer $L$.
\begin{equation}
\label{outputdnn}
    \mathbf{O}_t = \phi_o \left(\mathbf{H}_t^{(L)} \mathbf{W}_{h o}+\mathbf{b}_{o}\right)
\end{equation}

\section{Bidirectional Recurrent Neural Networks (BRNNs)}
Lets take an example of language modeling for now. Based on our current models we are able to reliably predict the next sequence element (i.e. the next word) based on what we have seen so far. However, there scenarios where we might want to fill in a gap in a sentence and the part of the sentence after the gap conveys significant information. This information is necessary to take into account to perform well on this kind of task \cite{zhang2019dive}. On a more generalised level we want to incorporate a look-ahead property for sequences.

\begin{figure}[H]
    \centering
    \includegraphics[width=0.8\textwidth]{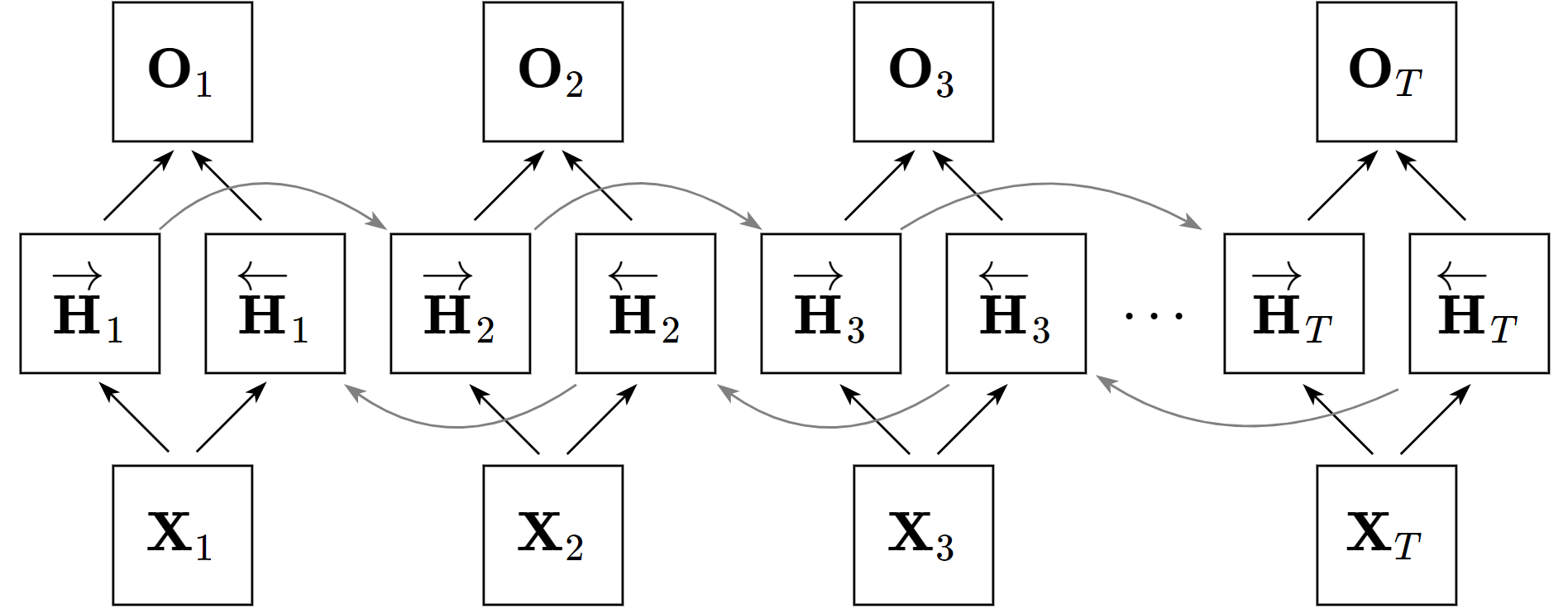}
    \caption{Architecture of a bidirectional recurrent neural network}
    \label{fig:brnn}
\end{figure}

To achieve this look-ahead property Bidirectional Recurrent Neural Networks (BRNNs) \cite{Schuster1997BidirectionalRN} got introduced which basically add another hidden layer which run the sequence backwards starting from the last element \cite{zhang2019dive}. An architectural overview can is visualised in Figure \ref{fig:brnn}. Here, we introduce a forward hidden state $\overrightarrow{\mathbf{H}}_{t} \in \mathbb{R}^{n \times h}$ and a backward hidden state $\overleftarrow{\mathbf{H}}_{t} \in \mathbb{R}^{n \times h}$. Their respective calculations are shown in Equation \ref{eq:forward} and Equation \ref{eq:backward}.
\begin{equation}
\label{eq:forward}
\overrightarrow{\mathbf{H}}_{t}=\phi\left(\mathbf{X}_{t} \mathbf{W}_{x h}^{(f)}+\overrightarrow{\mathbf{H}}_{t-1} \mathbf{W}_{h h}^{(f)}+\mathbf{b}_{h}^{(f)}\right)
\end{equation}
\begin{equation}
\label{eq:backward}
\overleftarrow{\mathbf{H}}_{t}=\phi\left(\mathbf{X}_{t} \mathbf{W}_{x h}^{(b)}+\overleftarrow{\mathbf{H}}_{t+1} \mathbf{W}_{h h}^{(b)}+\mathbf{b}_{h}^{(b)}\right)
\end{equation}

For that, we have similar weight matrices as in definitions before but now they are seperated into two sets. One set of weight matrices is for the forward hidden states $\mathbf{W}_{x h}^{(f)} \in \mathbb{R}^{d \times h}$ and $\mathbf{W}_{h h}^{(f)} \in \mathbb{R}^{h \times h}$ while the other one is for the backward hidden states $\mathbf{W}_{x h}^{(b)} \in \mathbb{R}^{d \times h}$ and $\mathbf{W}_{h h}^{(b)} \in \mathbb{R}^{h \times h}$. They also have their respective biases $\mathbf{b}_{h}^{(f)} \in \mathbb{R}^{1 \times h}$ and $\mathbf{b}_{h}^{(b)} \in \mathbb{R}^{1 \times h}$. With that, we can compute the output $\mathbf{O}_{t} \in \mathbb{R}^{n \times o}$ with $o$ being the number of outputs and $\frown$ denoting the concatenation of the two matrices on axis $0$ (stacking them on top of each other).
\begin{equation}
\mathbf{O}_{t}=\phi\left(\left[\overrightarrow{\mathbf{H}}_{t}^\frown \overleftarrow{\mathbf{H}}_{t}\right] \mathbf{W}_{h o}+\mathbf{b}_{o}\right)
\end{equation}

Again, we have weight matrices $\mathbf{W}_{h o} \in \mathbb{R}^{2 h \times o}$ and bias parameters $\mathbf{b}_{o} \in \mathbb{R}^{1 \times o}$. Keep in mind that the two directions can have different number of hidden units.

\section{Encoder-Decoder Architecture \& Sequence to Sequence (seq2seq)}
\label{seq2}
\begin{figure}[h]
    \centering
    \includegraphics[width=0.85\textwidth]{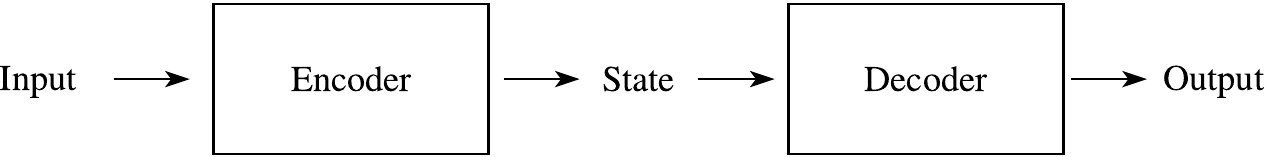}
    \caption{Encoder-Decoder Architecture Overview alternated from: \cite{zhang2019dive}}
    \label{fig:encoder-dec}
\end{figure}

The Encoder-Decoder architecture is a type of neural network architecture where the network is twofold. It consists of encoder network and a decoder network whose respective roles are to \textit{encode} the input into a state and \textit{decode} the state to an output. This state usually has shape of a vector or a tensor \cite{zhang2019dive}. A visualisation of this structure is shown in Figure \ref{fig:encoder-dec}.

Based on this Encoder-Decoder architecture a model called Sequence to Sequence (seq2seq) \cite{seq2seq} got proposed for generating a sequence output based on a sequence input. This model uses RNNs for the encoder as well as the decoder where the hidden state of the encoder gets passed to the hidden state of the decoder. Common applications of the model are Google Translate \cite{seq2seq, wu2016googles}, voice-enabled devices \cite{seq2seqspeech} or labeling video data \cite{venugopalan2015sequence}. It mainly focuses on mapping a fixed length input sequence of size $n$ to an fixed length output sequence of size $m$ where $n \neq m$ can be true but isn't a necessity.

\begin{figure}[h]
    \centering
    \includegraphics[width=0.8\textwidth]{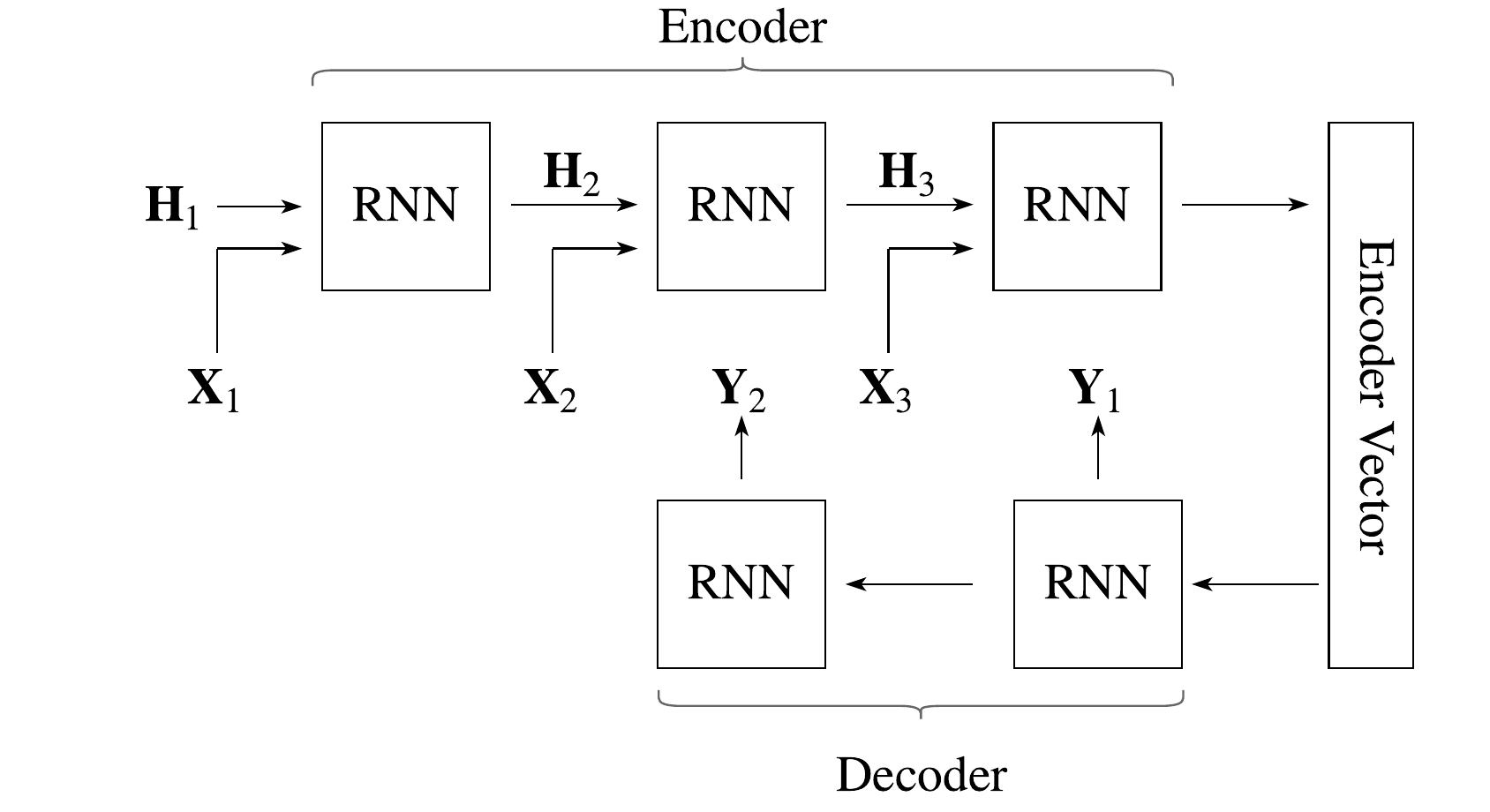}
    \caption{Visualisation of the Sequence to Sequence (seq2seq) Model}
    \label{fig:seq}
\end{figure}

A de-rellod visualisation of the proposed architecture is shown in Figure \ref{fig:seq}. Here, we have a encoder which consists of a RNN accepting a single element of the sequence $\mathbf{X}_t$ where $t$ is the order of the sequence element. These RNNs can be LSTMs or Gated Recurrent Units (GRUs) to further improve performance \cite{seq2seq}. Further, the hidden states $\mathbf{H}_t$ are computed according to the definition of the hidden states in the used RNN type (e.g. LSTM or GRU). The Encoder Vector (context) is a representation of the last hidden state of the encoder network which aims to aggregate all information from all previous input elements. This functions as initial hidden state of the decoder network of the model and enables the decoder to make accurate predictions. The decoder network again is built of a RNN which predicts an output $\mathbf{Y}_t$ at a time step $t$. The produced output is again a sequence where each $\mathbf{Y}_t$ is a sequence element with order $t$.  At each time step the RNN accepts a hidden state from the previous unit and itself produces an output as well as a new hidden state.

The Encoder Vector (context) was shown to be a bottleneck for these type of models since it needed to contain all all the necessary information of a source sentence in a fixed-length vector which was particularly problematic for long sequences. There have been approaches to solve this problem by introducing Attention in for example \cite{attention1} or \cite{attention2}. In the next section, we take a closer look at the proposed solutions.

\section{Attention Mechanism \& Transformer}

The Attention Mechanism for RNNs is partly motivated by human visual focus and the peripheral perception \cite{attention_blog}. It allows humans to focus on a certain region to achieve high resolution while adjacent objects are perceived with a rather low resolution. Based on these focus points and adjacent perception, we can make inference about what we expect to perceive when shifting our focus point. Similarly, we can transfer this method on our sequence of words where we are able to perform inference based on observed words. For example, if we perceive the word \textit{eating} in the sequence ``She is eating a green apple'' we assume to observe a food object in the near future \cite{attention_blog}.

\begin{figure}[h]
    \centering
    \includegraphics[width=0.35\textwidth]{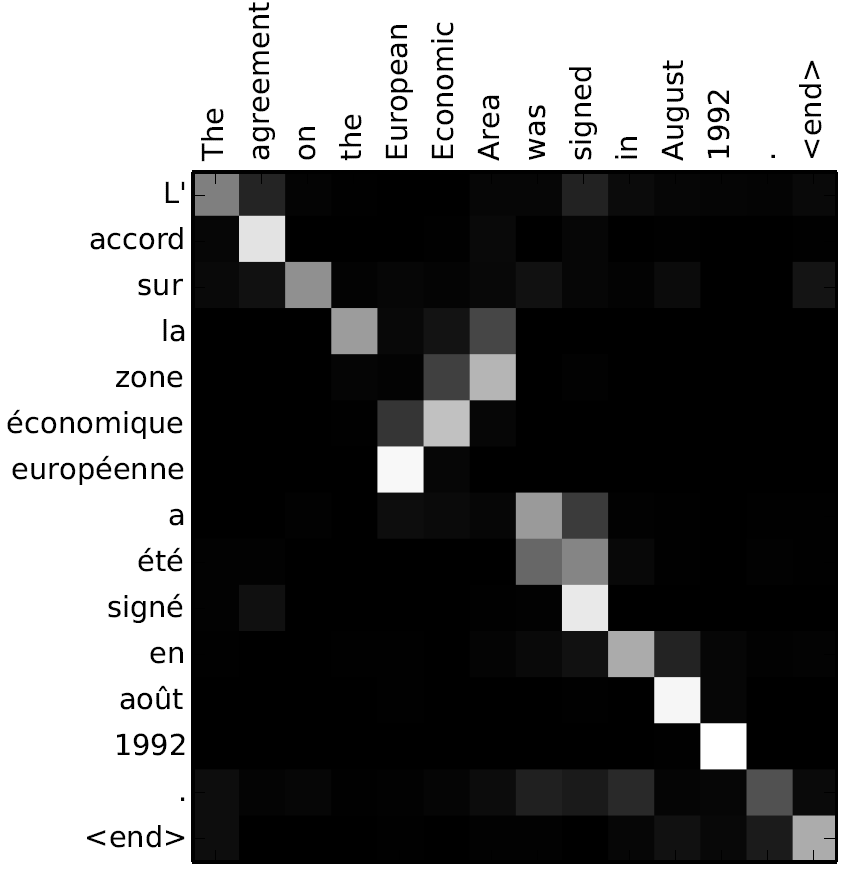}
    \caption{Example of an Alignment matrix of ``L'accord sur la zone \'{e}conomique europ\'{e}en a \'{e}t\'{e} sign\'{e} en ao\^{u}t 1992'' (French) and its English translation ``The agreement on the European Economic Area was signed in August 1992'': \cite{attention1}}
    \label{fig:matrix}
\end{figure}

Generally, Attention takes two sentences and transforms them into a matrix where each sequence element (i.e. a word) corresponds to a row or column. Based on this matrix layout we can fill in the entries to identify relevant context or correlations between them. An example of this process can be seen in Figure \ref{fig:matrix} where white denotes high correlation while black denotes low correlation. This method isn't limited to two sentences of a different languages as seen the example but can also be applied to the same sentence which is then called self-attention.

\subsection{Definition}

\begin{figure}[!ht]
    \centering
    \includegraphics[width=0.5\textwidth]{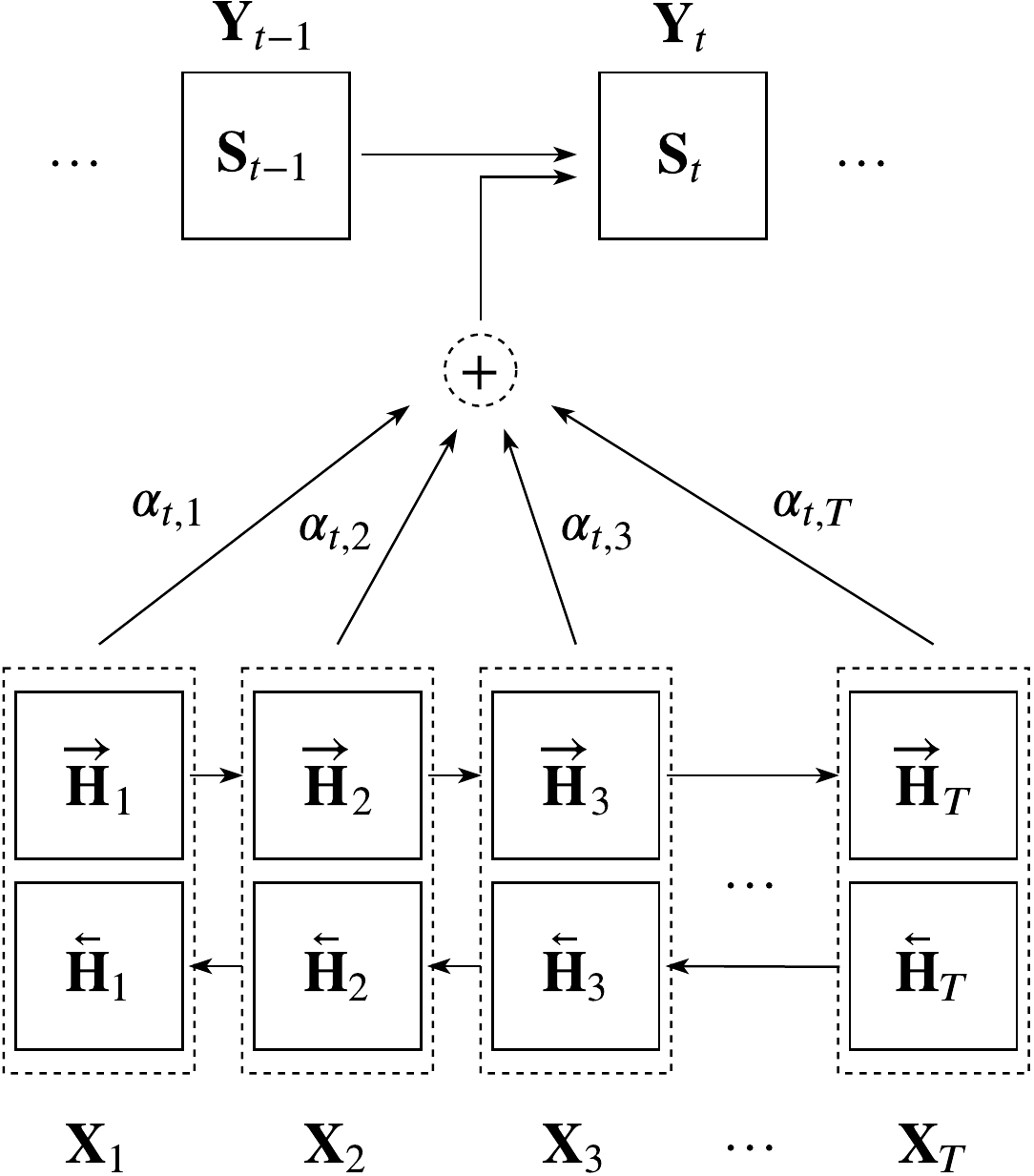}
    \caption{Encoder-Decoder architecture with additive attention mechanism alternated from: \cite{attention1}}
    \label{fig:attention_arch}
\end{figure}

To help the seq2seq model to better deal with long sequences the attention mechanism got introduced. Instead of constructing the Encoder Vector out of the last hidden state of the encoder network, attention introduces shortcuts between context vector and the entire source input. A visualisation of this process can be seen in Figure \ref{fig:attention_arch}. Here, we have source sequence $\mathbf{X}$ of length $n$ and try to output a target sequence $\mathbf{Y}$ of size $m$. In that regard the formulation is rather similar to the one we described before in Section \ref{seq2}. We have an overall hidden state $\mathbf{H}_{t^\prime}$ which is the concatenated version of the forward and backward pass as shown in Equation \ref{eq:concatenated}. Also, the hidden state of the decoder network is denoted as $\mathbf{S}_t$ while the encoder vector (context vector) is denoted as $\mathbf{C}_t$. Both of these are shown in Equation \ref{eq:decoder_state} and Equation \ref{eq:context_vector} respectively.
\begin{equation}
\label{eq:concatenated}
    \mathbf{H}_{t^\prime} = \left[\overrightarrow{\mathbf{H}}_{t^\prime}^\frown \overleftarrow{\mathbf{H}}_{t^\prime}\right]
\end{equation}
\begin{equation}
\label{eq:decoder_state}
    \mathbf{S}_t = \phi_d \left(\mathbf{S}_{t-1}, \mathbf{Y_{t-1}, \mathbf{C}_t} \right)
\end{equation}

The context vector $\mathbf{C}_t$ is a sum of hidden states of the input sequence each weighted with an alignment score $\alpha_{t, t^\prime}$ where $\sum_{t^\prime=1}^T \alpha_{t, t^\prime} = 1$. This is shown in Equation \ref{eq:context_vector} as well as Equation \ref{eq:alignment}.
\begin{equation}
\label{eq:context_vector}
    \mathbf{C}_t = \sum_{t^\prime=1}^T \alpha_{t, t^\prime} \cdot \mathbf{H}_{t^\prime}
\end{equation}
\begin{equation}
\label{eq:alignment}
    \alpha_{t, t^\prime} = \operatorname{align}(\mathbf{Y}_t, \mathbf{X}_{t^\prime}) = \frac{\operatorname{exp}\left(\operatorname{score}(\mathbf{S}_{t-1}, \mathbf{H}_{t^\prime})\right)}{\sum_{t^\prime = 1}^T\operatorname{exp}(\operatorname{score}(\mathbf{S}_{t-1}, \mathbf{H}_{t^\prime}))}
\end{equation}

The alignment  $\alpha_{t, t^\prime}$ connects an alignment score for the input at position $t^\prime$ and the output at position $t$. This score is based on how well this pair matches \cite{attention_blog}. The set of all alignment scores defines how much each source hidden state should be considered for each output \cite{attention_blog}. Please see Appendix \ref{app:seq2seq_attention} for a more easy and visual explanation of the attention mechanism in the seq2seq model.

\subsection{Different types of score functions}
Generally, there are different implementations for this score function which have been used in various works. Table \ref{tab:score} gives an overview over their respective name, equation and the usage in publications. Here, we have two trainable weight matrices in the alignment model denoted as $\mathbf{v}_a$ and $\mathbf{W}_a$.
\begin{table}[!ht]
    \centering
    \begin{tabular}{c|c|c}
        \toprule
         \textbf{Name} & \textbf{Equation for:} $\operatorname{score}(\mathbf{S}_t, \mathbf{H}_{t^\prime})$ & \textbf{Used In}\\
         \midrule
         Content-base &  $\operatorname{cosine}[\mathbf{S}_t, \mathbf{H}_{t^\prime}]$ & \cite{graves2014neural} \\
         \midrule
          Additive &  $\mathbf{v}_a^\top \tanh{\mathbf{W}_a[\mathbf{S}_t; \mathbf{H}_{t^\prime}]}$ & \cite{bahdanau2014neural} \\
          \midrule
          Location-Base & $\operatorname{softmax}(\mathbf{W}_a\mathbf{S}_t)$ & \cite{luong-etal-2015-effective} \\
          \midrule
          General & $\mathbf{S}_t^\top \mathbf{W}_a \mathbf{H}_{t^\prime}$ & \cite{luong-etal-2015-effective} \\
          \midrule
          Dot-Product & $\mathbf{S}_t^\top\mathbf{H}_{t^\prime}$ &  \cite{luong-etal-2015-effective} \\
          \midrule
          Scaled Dot-Product & $\frac{\mathbf{S}_t^\top\mathbf{H}_{t^\prime}}{\sqrt{n_{source}}}$  & \cite{attention} \\
        \bottomrule
    \end{tabular}
    \caption{Different score functions with their respective equations and usage alternated from: \cite{attention_blog}}
    \label{tab:score}
\end{table}

The Scaled-Dot-Product used in \cite{attention} scales the dot-product by the number of characters of the current word which is motivated by the problem that when the input is large, the softmax function may have an extremely small gradient which is a problem for efficient learning.

\subsection{Transformer}
\begin{figure}
    \centering
    \includegraphics[width=0.6\textwidth]{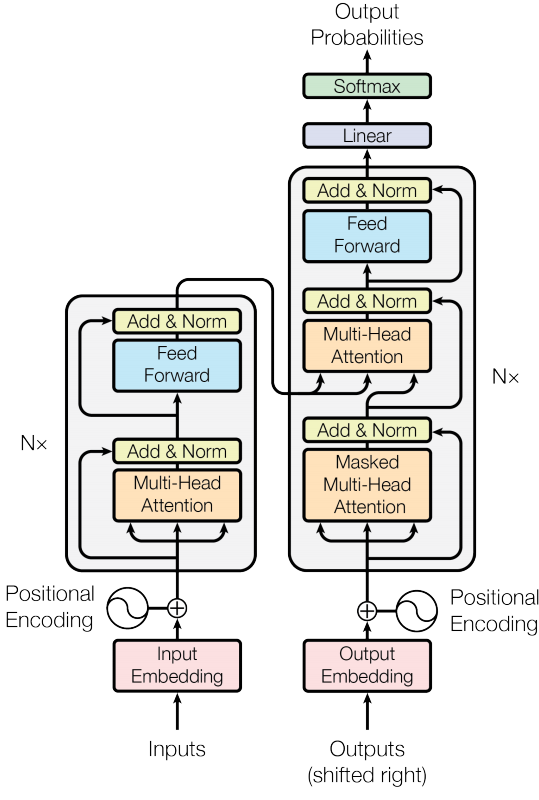}
    \caption{Model Architecture of the Transformer: \cite{attention}}
    \label{fig:transformer}
\end{figure}

By encorporating this Attention Mechanism the Transformer \cite{attention} got introduced which achieves parallelization by capturing recurrence sequence with attention but at the same time encoding each item's position in the sequence based on the encoder-decoder architecture \cite{zhang2019dive}. In fact, for that it doesn't use any recurrent network units and entirely relies on the self-attention mechanism to improve performance. The encoding part of the architecture is made out of several encoders (e.g. six encoders in \cite{attention}) while the decoder part consists out of decoders with the same amount as the encoders. A general overview over the architecture is illustrated in Figure \ref{fig:transformer}.

Here, each encoder component consists out of two sub-layers which are Self-Attention and a Feed Forward Neural Network. Similarly, those two sub-layers are found in each decoder component but with a Encoder-Decoder Attention sub-layer in between them which works similarly to the Attention used in the seq2seq model. The deployed Attention layers are not your ordinary attention layers but a method called Multi-Headed Attention which improves performance of the attention layer. This allows the model to jointly attend to information from different representation
subspaces at different positions which in easier terms runs different chunks in parallel and concatenates the results \cite{attention}. Unfortunately, explaining the design choices and mathematical formulations contained in multi-headed attention would be to much details at this point. Please refer to the original paper \cite{attention} for more information. The architecture shown in Figure \ref{fig:transformer} also deploys skip connections and layer normalisation for each sub-layer of the encoder as well as the decoder. One thing to note is that the input as well as the output get embedded and a positional encoding is applied which represents the proximity of sequence elements (see Appendix \ref{app:pos}).

The final linear and softmax layer turn the vector of floats which is the output of the decoder stack into a word. This is done by transforming the vector through the linear layer into a much larger vector called a logits vector \cite{transformer_blog}. This logits vector has the size of the learned vocabulary from the training dataset where each cell corresponds to the score of a unique word \cite{transformer_blog}. By applying a softmax function we turn those scores into probabilities which sum up to $1$ and therefore we can choose the cell (i.e. the word) with the highest probability as output for this particular time step.

\section{Pointer Networks (Ptr-Nets)}
Pointer Networks (Ptr-Nets) \cite{pointer} adapt the seq2seq model with attention to improve it by not fixing the discrete categories (i.e. elements) of the output dictionary \textit{a priori}. Instead of yielding an output sequence generated from an input sequence, a pointer network creates a succession of pointers to the elements of the input series \cite{pointer_blog}. In \cite{pointer} they show that by using Pointer Networks they can solve combinatorial optimization problems such as computing planar convex hulls, Delaunay triangulations and the symmetric planar Travelling Salesman Problem (TSP).

Generally, we apply additive attention (from Table \ref{tab:score}) between states and then normalize it by applying the softmax function to model the output conditional probability as seen in Equation \ref{eq:pointer}.
\begin{equation}
\label{eq:pointer}
    \mathbf{Y}_t = \operatorname{softmax}\left(\operatorname{score}\left(\mathbf{S}_t, \mathbf{H}_{t^\prime}\right)\right) = \operatorname{softmax}\left( \mathbf{v}_a^\top \tanh{\mathbf{W}_a[\mathbf{S}_t; \mathbf{H}_{t^\prime}]}\right)
\end{equation} 

The attention mechanism is simplified, as Ptr-Net does not blend the encoder states into the output with attention weights. In this way, the output only responds to the positions but not the input content \cite{attention_blog}.

\section{Conlusion \& Outlook}
In this work we gave an introduction into fundamentals for Recurrent Neural Networks (RNNs). This includes the general framework for RNNs, Backpropagation through time, problems of traditional RNNs, LSTMS, Deep and Bidirectional RNNs as well as more recent advances such as the Encoder-Decoder Architecture, seq2seq model, Attention, Transformer and Pointer Networks. Most topics are only covered conceptionally and don't go too deep into implementation specifications. To get a broader understanding of the covered topics, we recommend looking into some of the cited original papers. Additionally, most recent publications use some of the presented concepts so we recommend taking a look at such papers.

One recent publication which uses many of the presented concepts is ``Grandmaster level in StarCraft II using multi-agent reinforcement learning'' by Vinyals et al. \cite{Vinyals2019}. Here, they present their approach to train agents to play the real-time strategy game Starcraft II with great success. If the presented concepts were a little too theoretical for you we recommend reading that paper to see LSTMs, the Transformer or Pointer Networks in a setting which can be deployed in a more practical environment. 

\printbibliography[]

\begin{appendices}
\section{Visual Representation of LSTMs}
\label{app:vis}
In this section we consecutively construct the full architecture of Long Short-Term Memory Units (LSTMs) explained in Section \ref{sec:lstms}. For a description what is changing between each step please read Section \ref{sec:lstms} or refer to the source of the illustrations \cite{zhang2019dive}.

\begin{figure}[H]
    \centering
    \includegraphics[height=6cm]{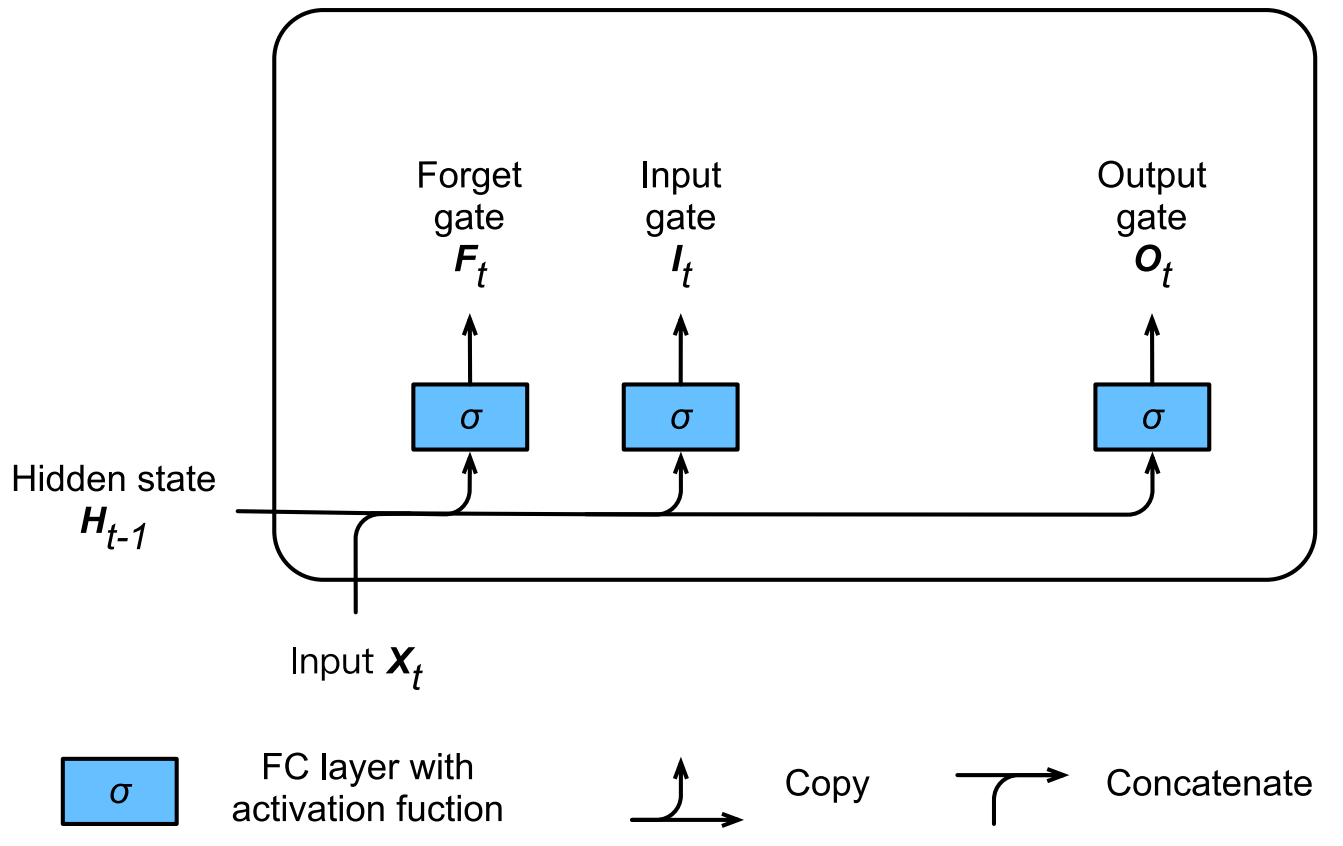}
    \caption{Calculation of input, forget, and output gates in an LSTM: \cite{zhang2019dive}}
    \label{fig:gates}
\end{figure}

\begin{figure}[H]
    \centering
    \includegraphics[height=6cm]{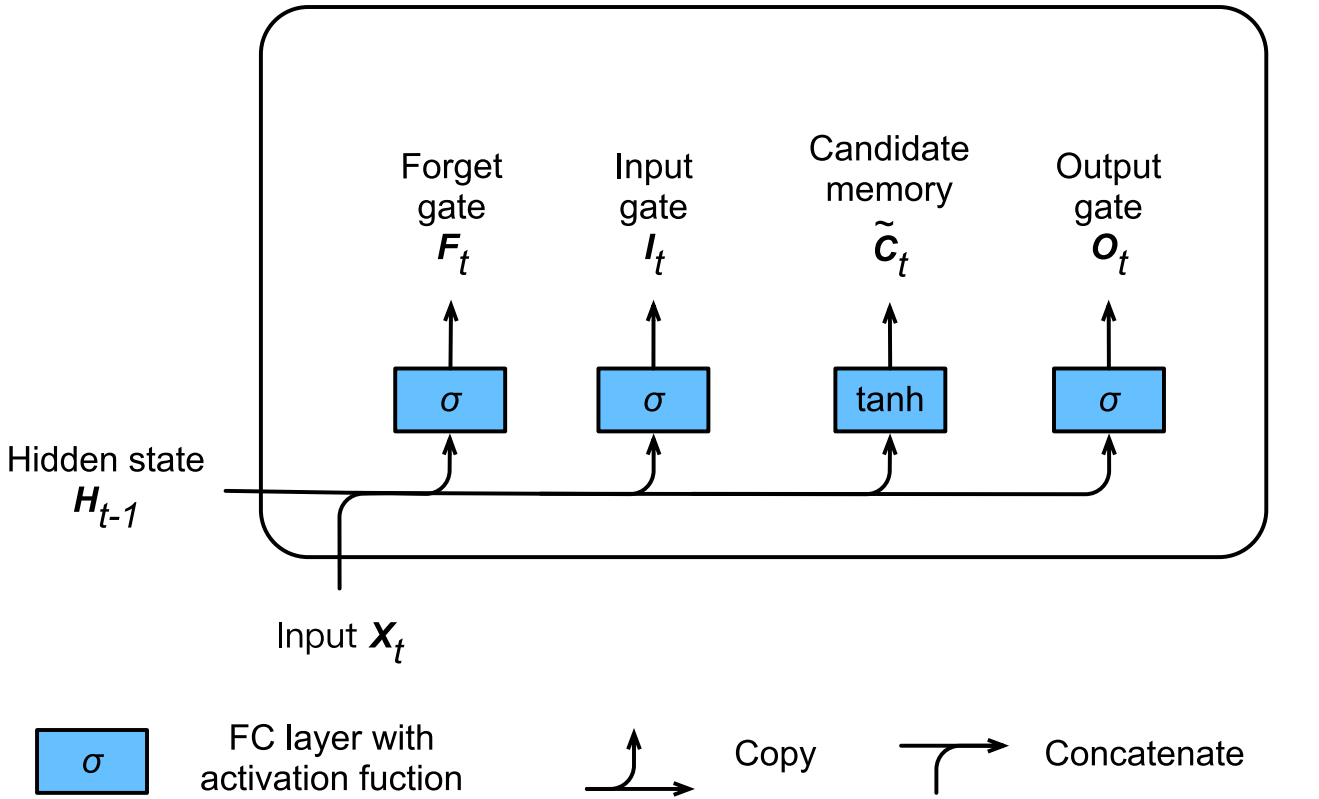}
    \caption{Computation of candidate memory cells in LSTM: \cite{zhang2019dive}}
    \label{fig:candidate}
\end{figure}

\begin{figure}[H]
    \centering
    \includegraphics[height=6cm]{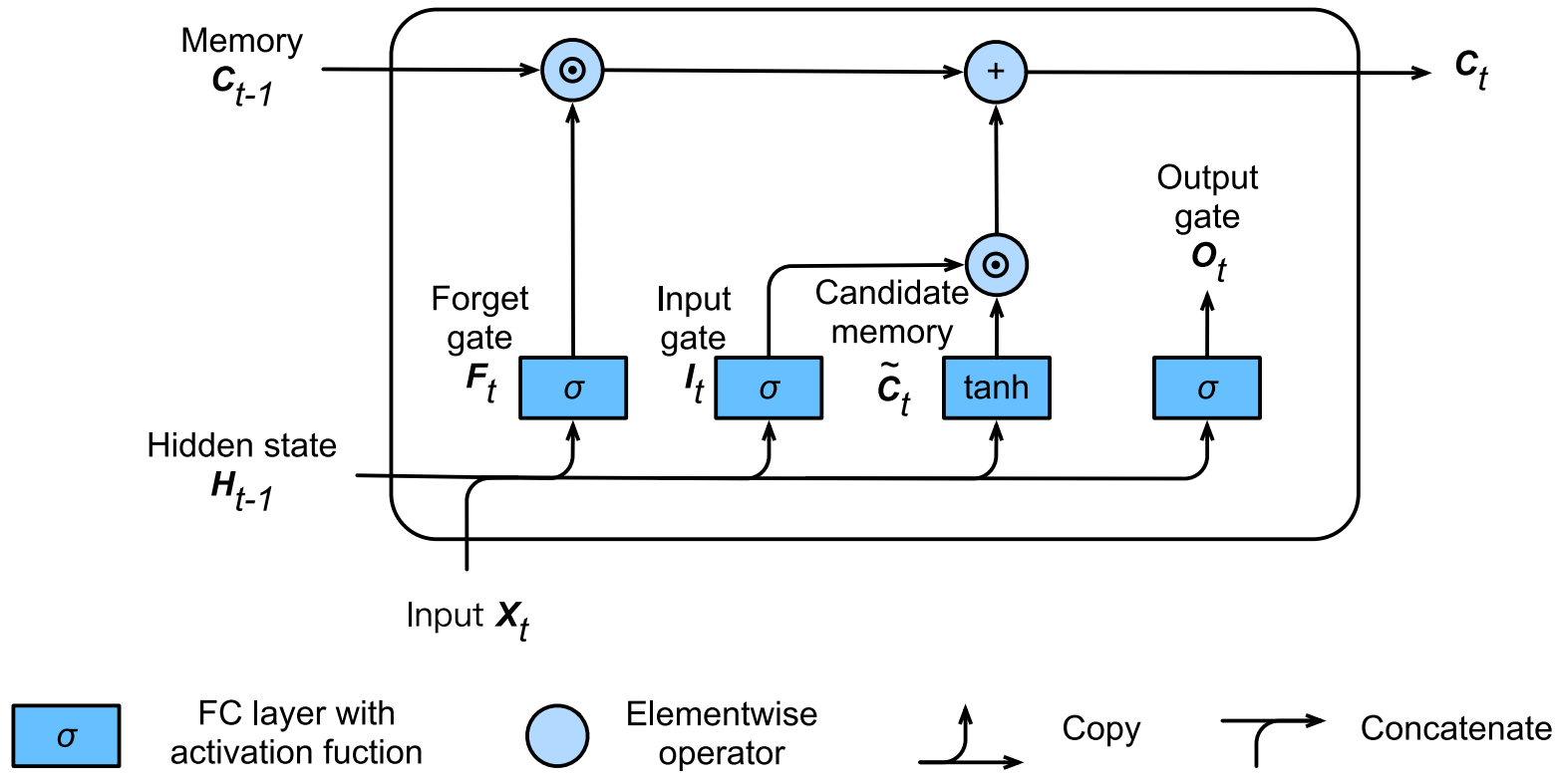}
    \caption{Computation of memory cells in an LSTM: \cite{zhang2019dive}}
    \label{fig:memory}
\end{figure}

\begin{figure}[H]
    \centering
    \includegraphics[height=6cm]{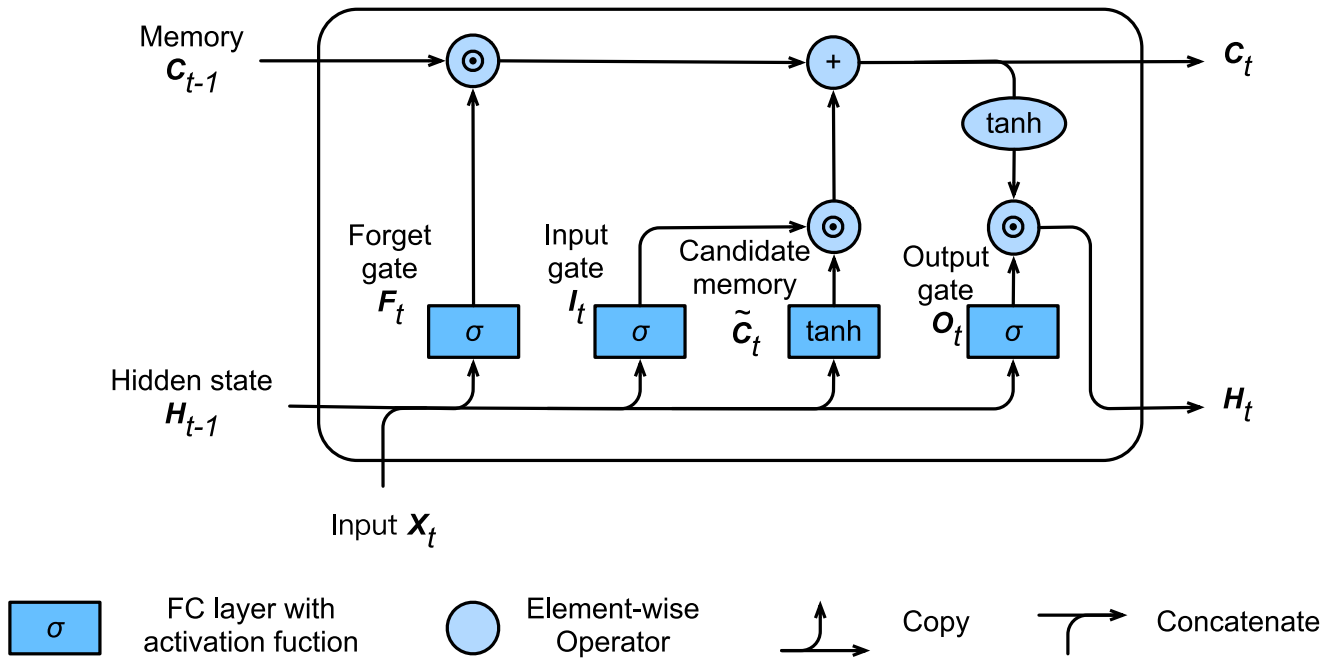}
    \caption{Computation of the hidden state in an LSTM: \cite{zhang2019dive}}
    \label{fig:hidden}
\end{figure}

\section{Visual Representation of seq2seq with Attention}
\label{app:seq2seq_attention}
\begin{figure}[H]
        \centering
        \includegraphics[width=\textwidth, height=6.3cm]{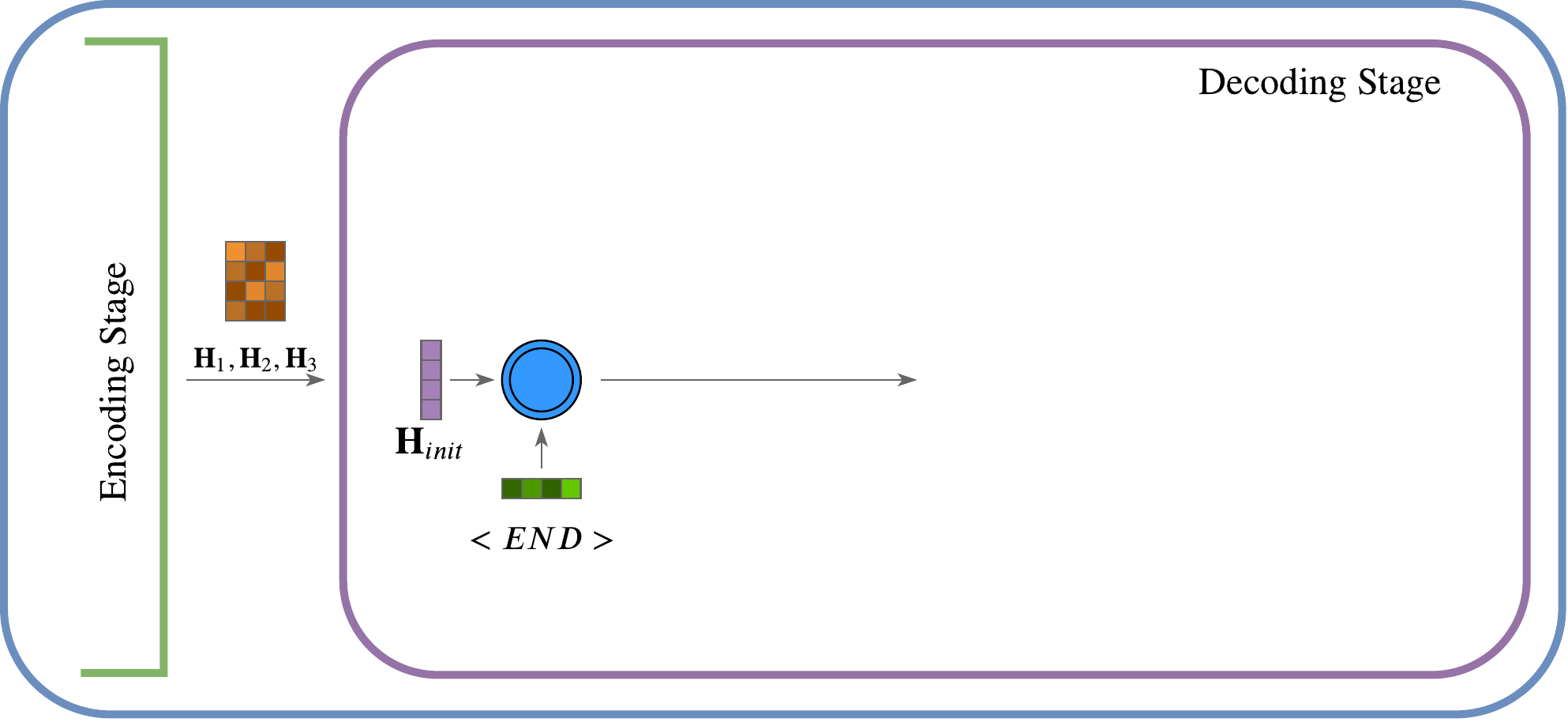}
        \caption{Seq2Seq Model with Attention Mechanism Step 1 alternated from: \cite{seq2seqBlog}}
        \label{fig:seq2seqframe1}
    \end{figure}

The seq2seq model with attention passes a lot more data from the encoder to the decoder than the regular seq2seq model. Instead of passing the last hidden state of the encoding stage, the encoder passes all the hidden states to the decoder. The first step of the decoder part in the seq2seq model with attention is illustrated in Figure \ref{fig:seq2seqframe1} where we pass ``\textit{I am a student}'' to the encoder and expect a translation to french producing ``\textit{je suis un \'{e}tudiant}''. Here, all the hidden states of the encoder $\mathbf{H}_1$, $\mathbf{H}_2$, $\mathbf{H}_3$ are passed to the attention decoder as well as the embedding from the $<End>$ token and an initial decoder hidden state $\mathbf{H}_{init}$.

\begin{figure}[H]
        \centering
        \includegraphics[width=\textwidth, height=6.3cm]{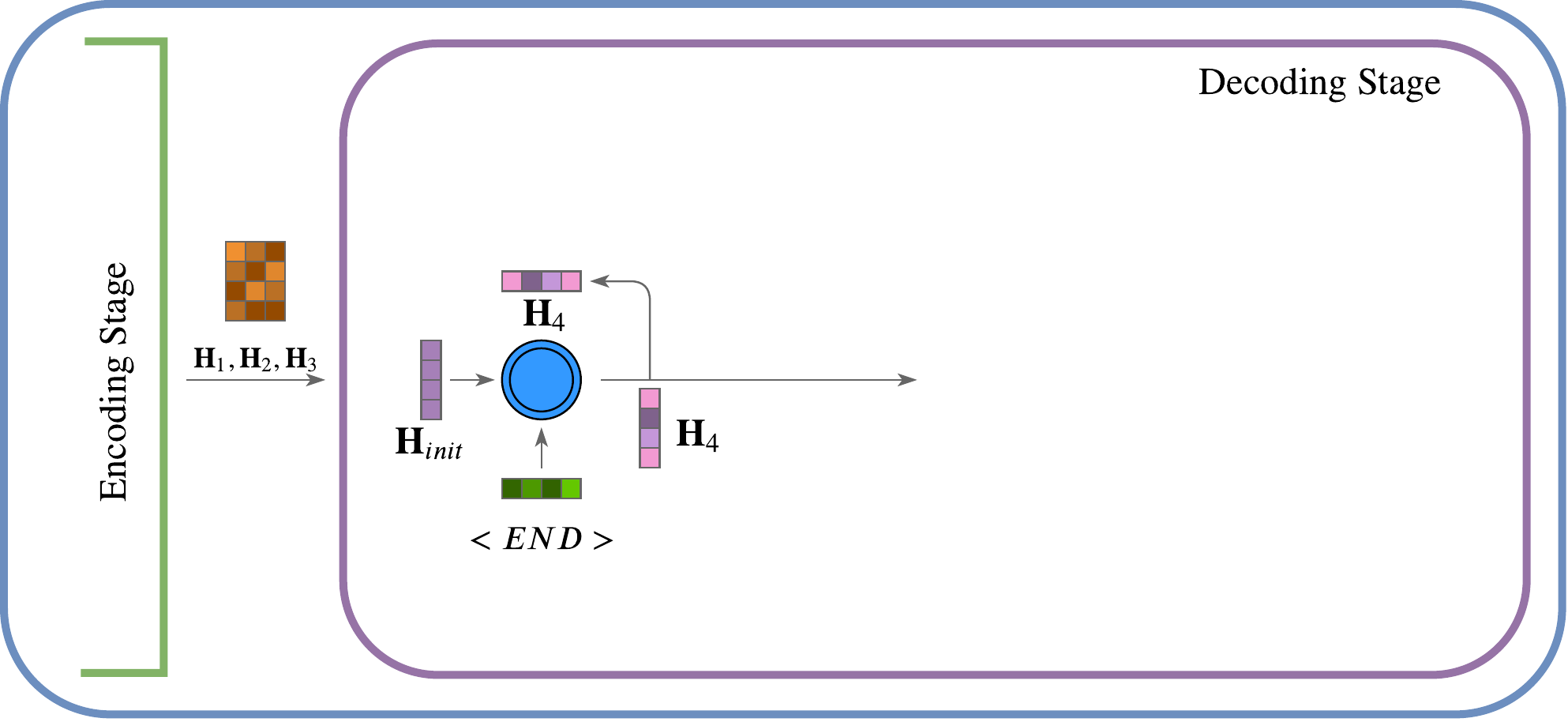}
        \caption{Seq2Seq Model with Attention Mechanism Step 2 alternated from: \cite{seq2seqBlog}}
        \label{fig:seq2seqframe2}
    \end{figure}

Next, we produce an output and a new hidden state vector $\mathbf{H}_4$. However, the output is discarded. This can be seen in Figure \ref{fig:seq2seqframe2}.

\begin{figure}[H]
        \centering
        \includegraphics[width=\textwidth, height=6.3cm]{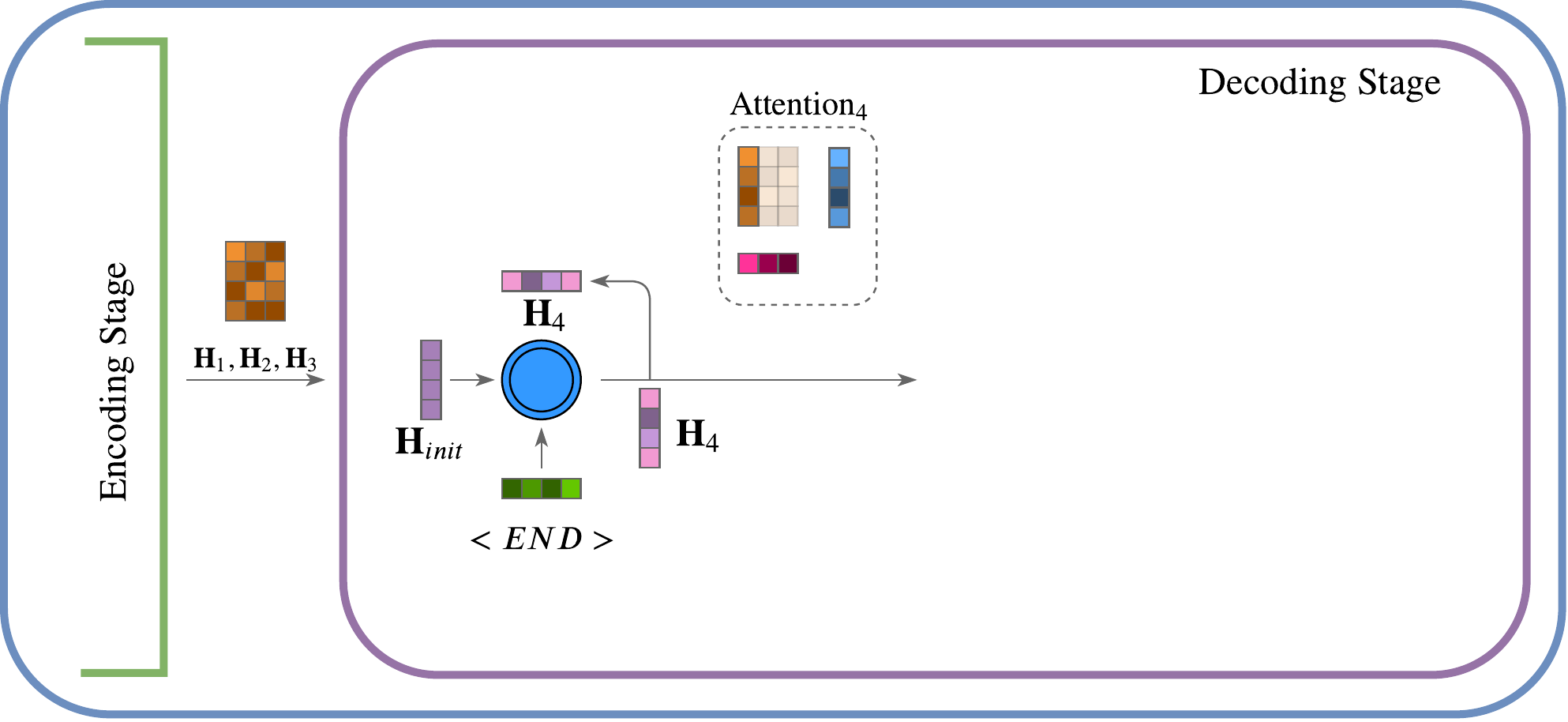}
        \caption{Seq2Seq Model with Attention Mechanism Step 4 alternated from: \cite{seq2seqBlog}}
        \label{fig:seq2seqframe4}
    \end{figure}

For the attention step we use this produced hidden state vector $\mathbf{H}_4$ and the hidden states from the encoder $\mathbf{H}_1$, $\mathbf{H}_2$, $\mathbf{H}_3$ to produce a context vector $\mathbf{C}_4$ (blue). This process can be seen in Figure \ref{fig:seq2seqframe4}. Each encoder hidden state is most associated with a certain word in the input sentence \cite{seq2seqBlog}. When we give these hidden states scores and apply a softmax to it we generate probability values. These probabilities are represented with by the three-element pink vector where light values stand for high probabilities while dark values denote low probabilities. Next, we apply each hidden state vector $\mathbf{H}_1$, $\mathbf{H}_2$, $\mathbf{H}_3$ by its softmaxed score which increases hidden states with high scores, and decreases hidden states with low scores. This is visualised by graying out the hidden states $\mathbf{H}_2$ and $\mathbf{H}_3$ while keeping $\mathbf{H}_1$ in solid color.

\begin{figure}[H]
        \centering
        \includegraphics[width=\textwidth, height=6.3cm]{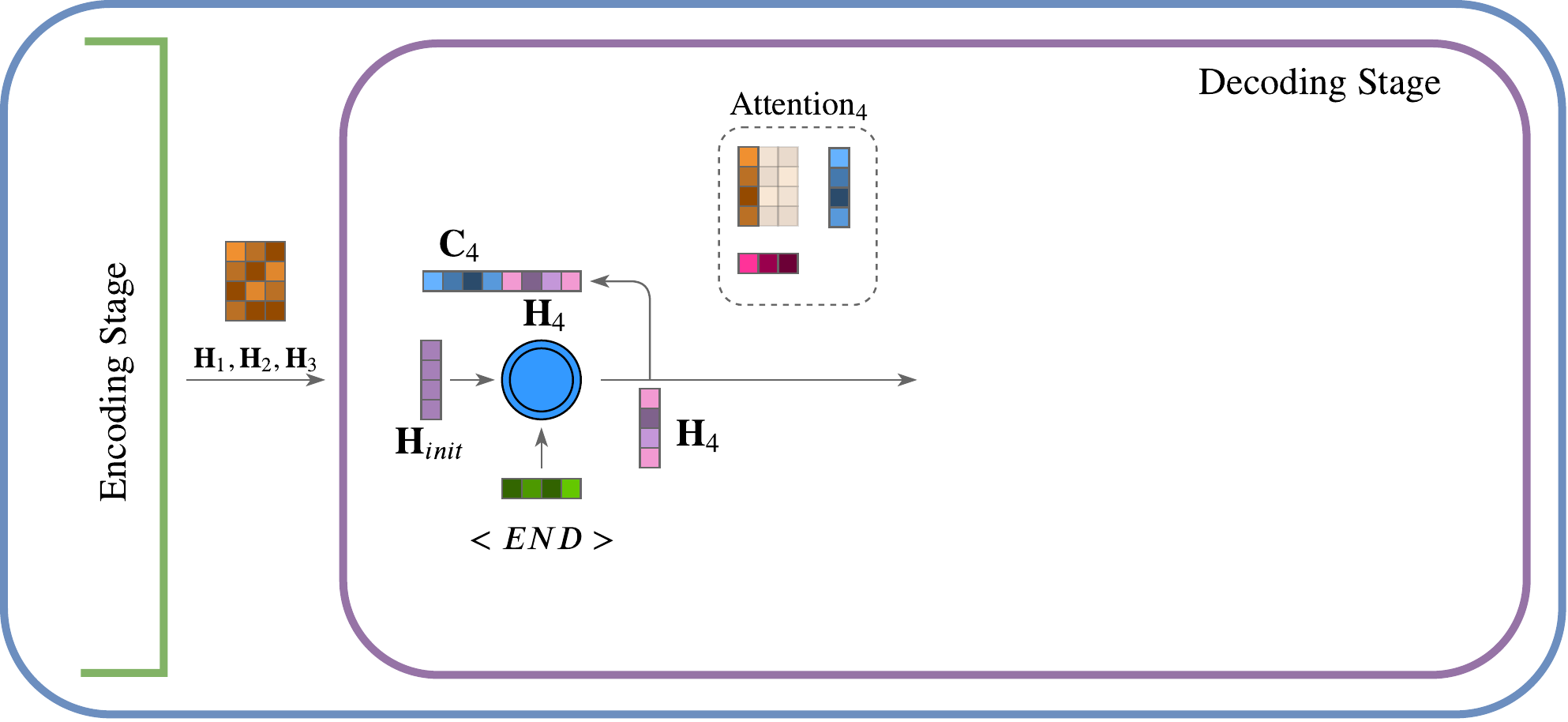}
        \caption{Seq2Seq Model with Attention Mechanism Step 5 alternated from: \cite{seq2seqBlog}}
        \label{fig:seq2seqframe5}
    \end{figure}

After that, we concatenate this produced context vector $\mathbf{C}_4$ with the produced hidden state $\mathbf{H}_4$. One can see this process in Figure \ref{fig:seq2seqframe5}. This process just stacks the two vectors on top of each other.

\begin{figure}[H]
        \centering
        \includegraphics[width=\textwidth, height=6.3cm]{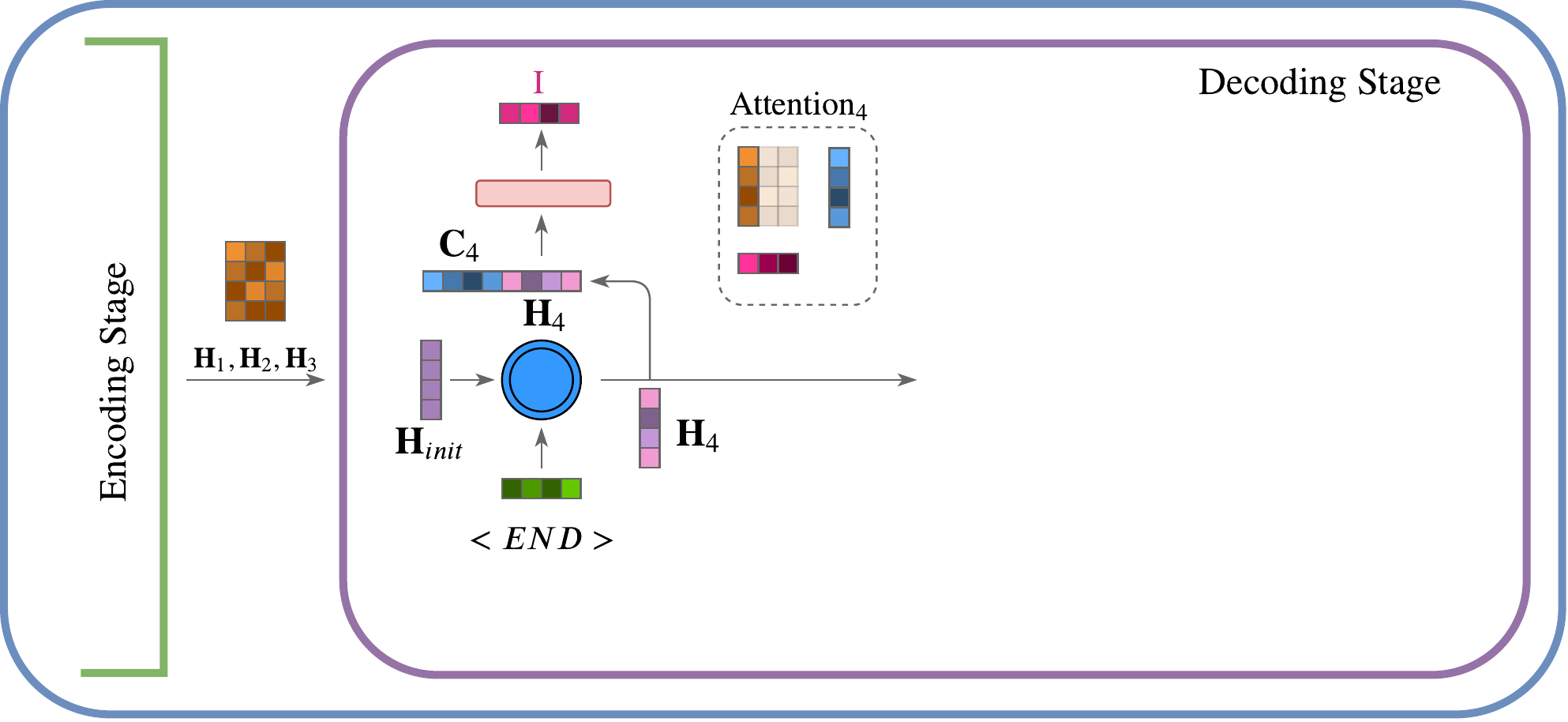}
        \caption{Seq2Seq Model with Attention Mechanism Step 6 alternated from: \cite{seq2seqBlog}}
        \label{fig:seq2seqframe6}
    \end{figure}

This concatenated version of hidden state $\mathbf{H}_4$ and context vector $\mathbf{C}_4$ is then passed into a jointly trained Feedforward Neural Network. This network is visualised by the red box with round edges in Figure \ref{fig:seq2seqframe6}. The output of this network then represents the output of the current time step $t$ which in this case represents the word ``\textit{I}''. This basically concludes all the steps needed at each iteration step.

\begin{figure}[H]
        \centering
        \includegraphics[width=\textwidth, height=6.3cm]{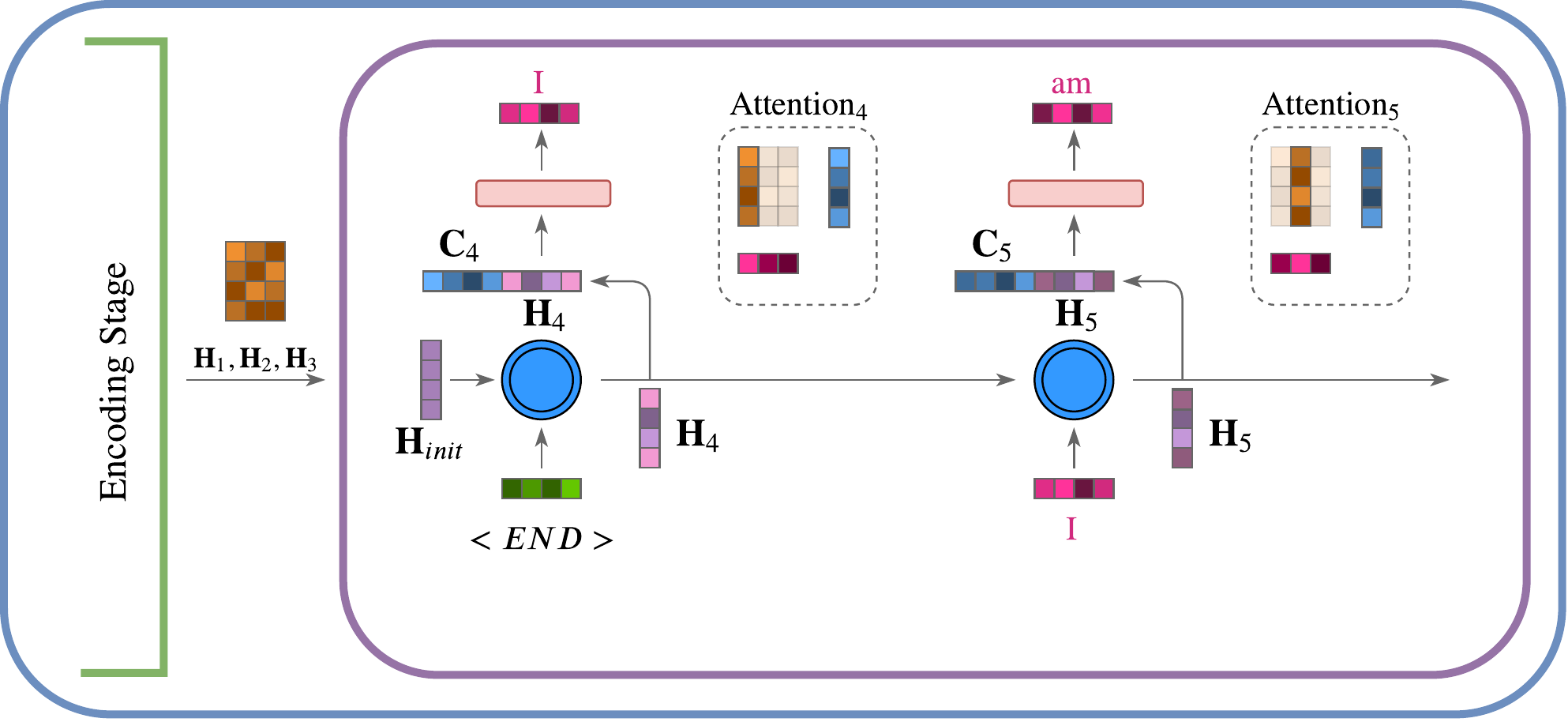}
        \caption{Seq2Seq Model with Attention Mechanism Step 7 alternated from: \cite{seq2seqBlog}}
        \label{fig:seq2seqframe7}
    \end{figure}

If we take a look at the next iteration step in Figure \ref{fig:seq2seqframe7} we can see that the output from the previous hidden state $\mathbf{H}_4$ is passed instead of the $<END>$ token. All the other steps are equal from the previous iteration. However, we can see that the hidden state $\mathbf{H}_2$ has the best score during the attention stage. Again, this is represented by the lightest shade of pink in the score vector. By multiplying the scores with the hidden states we achieve two reduced hidden states $\mathbf{H}_1$ and $\mathbf{H}_3$ while keeping $\mathbf{H}_2$ as the most active hidden state. This results in the word ``\textit{am}'' being produced as the output of the Feedforward Neural Network for this time step.

Obviously, there are still two more attention decoder time steps which are omitted here for illustration purposes. The functionality of each of those steps however would still be equivalent to the already seen time steps.

\section{Visual Representation of Positional Encodings used in the Transformer}
\label{app:pos}
\begin{figure}
    \centering
    \includegraphics[width=\textwidth]{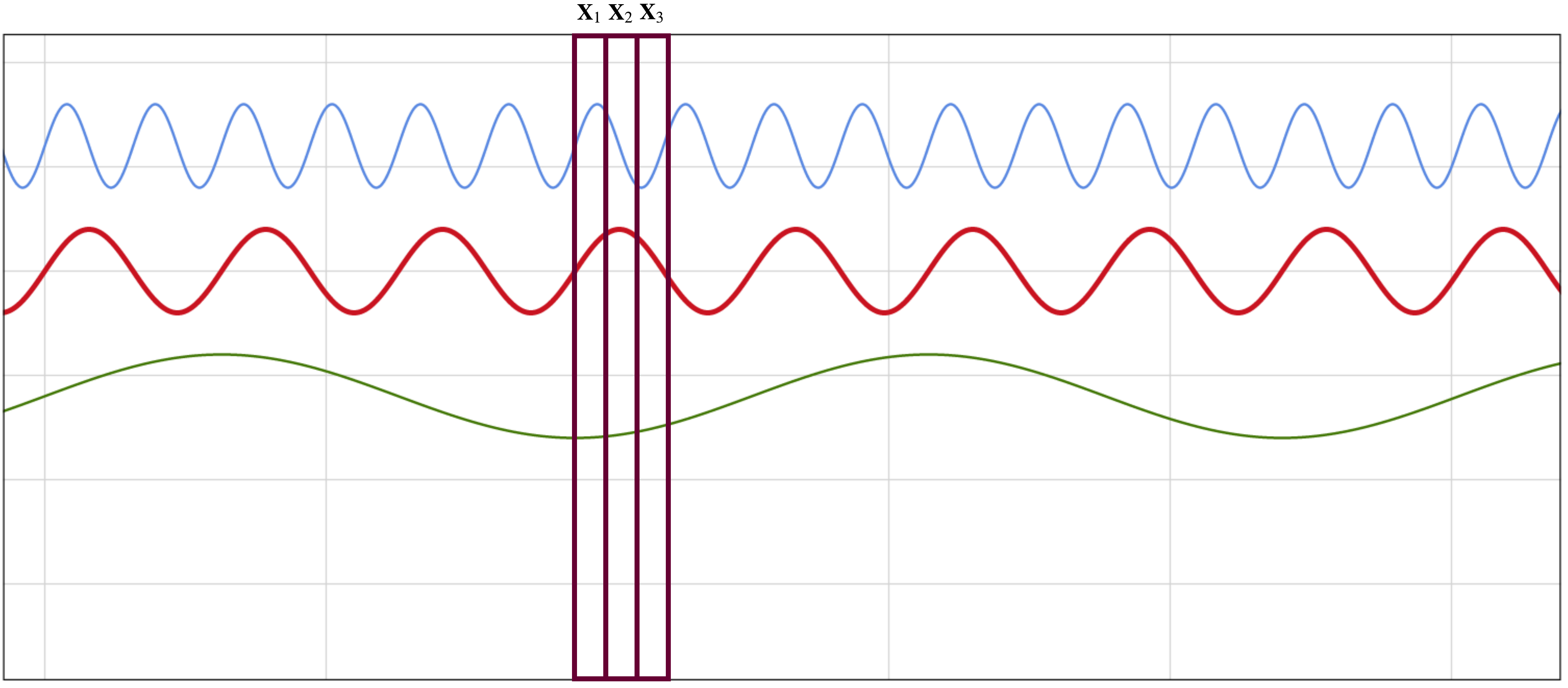}
    \caption{Positional Encoding Example based on trigonometric functions}
    \label{fig:pos_encoding}
\end{figure}

On example of a positional encoding used inside the transformer is applying trigonometric functions as seen in Figure \ref{fig:pos_encoding}. Here, we have multiple trigonometric functions with different frequency. We also show the encoding for three words i.e. $\mathbf{X}_1$, $\mathbf{X}_2$, $\mathbf{X}_3$. 

In principal the encoding for $\mathbf{X}_1$ is therefore high for the first curve (blue), mid for the second curve (red) and low for the last curve (green). Similarly, this applies for the other words as well. What we can see here is that close words have closer encodings while distant words have more different encodings. Generally, this is a method for binary encoding the position of a given sequence. 

The choice of such a positional encoding algorithm definitely is not the main contribution of \cite{attention} but it is a relevant concept to at least understand in theory since this boosts performance.

\end{appendices}

\end{document}